%% file: emnlp2018.tex
\documentclass[11pt,a4paper]{article}
\usepackage[hyperref]{emnlp2018}
\usepackage{times}
\usepackage{latexsym}
\usepackage{amsfonts}
\usepackage{url}
\usepackage{amsmath}
\usepackage{graphicx}
\usepackage{booktabs}
\usepackage{microtype}
\usepackage{subcaption}
\usepackage{tikz}
\usetikzlibrary{matrix,decorations.pathreplacing, calc, positioning}
\graphicspath{ {./img/} }
 
\aclfinalcopy 

\title{Iterative Recursive Attention Model for\\Interpretable Sequence Classification}

\author{Martin Tutek \and Jan \v{S}najder\\
Text Analysis and Knowledge Engineering Lab\\
Faculty of Electrical Engineering and Computing, University of Zagreb\\
Unska 3, 10000 Zagreb, Croatia \\
\tt \{martin.tutek,jan.snajder\}@fer.hr
}

\date{}

\begin{document}
\maketitle
\begin{abstract}

Natural language processing has greatly benefited from the introduction of the
attention mechanism. However, standard attention models are of limited
interpretability for tasks that involve a series of inference steps. We
describe an iterative recursive attention model, which constructs 
incremental representations of input data through reusing results of previously
computed queries. We train our model on sentiment classification
datasets and demonstrate its capacity to identify and combine different aspects
of the input in an easily interpretable manner, while obtaining performance
close to the state of the art.

\end{abstract}

\input{1-Introduction}

\input{2-RelatedWork}

\input{3-Methodology}
\input{4-Experiments}

\input{5-Conclusion}

\section*{Acknowledgment} This research has been supported by the European Regional Development Fund under the grant KK.01.1.1.01.0009 (DATACROSS).

\bibliography{emnlp2018}
\bibliographystyle{acl_natbib_nourl}

\end{document}

%% file: 1-Introduction.tex
\section{Introduction}
\label{sec:introduction}

The introduction of the attention mechanism \cite{bahdanau2014neural} offered a
way to demystify the inference process of neural models.  By assigning scalar
weights to different elements of the input, we are able to visualize and
potentially understand why the model made the decision it made, or discover a
deficiency in the model by tracing down a relevant aspect of the input being
overlooked by the model. Specifically in natural language processing (NLP),
which abounds with variable-length word sequence classification tasks, attention
alleviates the issue of learning long-term dependencies in recurrent neural
networks \cite{bengio1994learning} by offering the model a glimpse into
previously processed tokens. 

Attention offers a good retrospective explanation of the classification decision 
by indicating what parts of the input contributed the most to the 
decision. However, in many cases the final decision is best interpreted as a
result of a series of inference steps, each of which can potentially affect its 
polarity. A case in point is sentiment analysis, in which
contrastive clauses and negations act as polarity switches of the overall
sentence sentiment. In such cases, attention will only point to the part of the
input sentence whose polarity matches that of the final decision.  However,
unfolding the inference process of a model into a series of interpretable steps
would make the model more interpretable and allow one to identify its
shortcomings.

As a step toward that goal, we propose an extension of the iterative attention
mechanism \cite{sordoni2016iterative}, which we call the \emph{iterative recursive attention model} (IRAM), where the result of an attentive query is nonlinearly
transformed and then added to the set of vector representations of the input
sequence. The nonlinear transformation, along with reusing the representations
obtained in previous steps, allows the model to construct a recursive 
representation and process the input sequence bit by bit.  The upshot is
that we can inspect how the model weighs the different parts
of the sentence and recursively combines them to give the final
decision.  We test the model on two sentiment analysis tasks and demonstrate
its capacity to isolate different task-related aspects of the input, while 
reaching performance comparable with the state of the art.

%% file: 2-RelatedWork.tex
\section{Related Work}
\label{sec:rw}

Attention \cite{bahdanau2014neural} and its variants 
\cite{luong2015effective} have initially been proposed for machine translation,
but are now widely adopted in NLP. Attention has proven especially
useful in tasks that involve long text sequences, such as
summarization \cite{rush2015neural, see2017get}, question answering
\cite{hermann2015teaching, xiong2016dynamic, cui2016attention}, and natural
language inference \cite{rocktaschel2015reasoning, yin2015abcnn,
parikh2016decomposable}, as well as purely attentional machine translation \cite{vaswani2017attention, gu2017non}. 

Thus far, there has been a number of interesting and effective approaches for interpreting the inner workings of 
recurrent neural networks through methods such as representing them as finite automata \cite{weiss2017extracting}, extracting
inference rules \cite{zanzotto2017can}, and analyzing saliency of inputs through first-order derivative 
information \cite{li2016visualizing, arras2017explaining}. 

Akin to the saliency analysis approaches, we opt not to condense the 
trained network into a finite set of rules. We differ from \cite{li2016visualizing, arras2017explaining} in that we attempt
to decode the steps of the decision process of a recurrent network instead of demonstrating through saliency how the decision
changes with respect to the inputs. In the context of sentiment analysis, the main benefit we see in representing the decision process of a recurrent network as
a sequence of steps is that it offers a simple way to isolate sentiment-bearing phrases by observing how they get grouped in a single iteration.
Secondly, we aim for improved interpretability of functional dependencies such as negation, where we demonstrate that our method first attends on
the negated phrase, constructing an intermediate representation, which is then recursively transformed in the next iteration.

\newcite{sordoni2016iterative} introduced the iterative attention mechanism for 
question answering, where attention alternates between the question 
and the document, and the query is updated in 
each step by a GRU cell \cite{cho2014properties}. The model combines the
weights obtained throughout the iterations to select the final answer,
similar to the attention sum reader of \newcite{kadlec2016text} and pointer networks of \newcite{vinyals2015pointer}.

We believe there is much to gain from the iterative attention mechanism by
eliminating the direct link between the intermediate representations and the
output, allowing the model to construct its own sequential representation of
the input. Our model only connects the last attention step 
 to the output, removing the need for intermediate steps
to contain all the information relevant for the final decision. Apart from
\cite{sordoni2016iterative}, related work closest to ours consists of concepts of multi-head
attention \cite{lin2017structured, vaswani2017attention}, in which
all queries are generated at once, pairwise attention
\cite{cui2016attention,xiong2016dynamic}, where attention is applied to multiple
inputs but is not applied iteratively and hierarchical iterative attention \cite{yang2016hierarchical}, 
where the authors first use a intra-sentence attention mechanism and then combine the intermediate representations
with inter-sentence attention. In contrast to their work, we do not predetermine the level on which the attention is applied --
in each iteration the mechanism can focus on any element of the input sequence.

%% file: 3-Methodology.tex
\section{Model}
\label{sec:model}

Throughout the experiments, we will use two variants of our model: (1) the vanilla model and (2) the full model. The vanilla model 
contains the bare minimum of components needed for the attention mechanism to function as intended. The purpose of the vanilla model 
is to eliminate any additional confounders for the performance and showcase the interpretability of the model. For the full model, we extend the vanilla model with additional deep learning components commonly employed in state-of-the-art models, to showcase the performance of the model when given
capacity akin to competing models.

In both versions of the model, data is processed in three phases: 
(1) \emph{encoding phase}, which contextualizes the word representations; (2) \emph{attention
phase}, which uses iterative recursive attention to
isolate and combine the different parts of the input; and (3) \emph{classification
phase}, where the learned representation is fed as an input to a classifier.

The vanilla and full model differ only in the encoding phase, while our proposed attention mechanism is employed 
only in the second phase. We begin with a detailed account of the proposed attention mechanism and 
its regularization, and continue with a description of the remaining components, highlighting the differences between the vanilla and
full models.

\subsection{Iterative Recursive Attention} 

\begin{figure} 
\centering
\includegraphics[width=6cm]{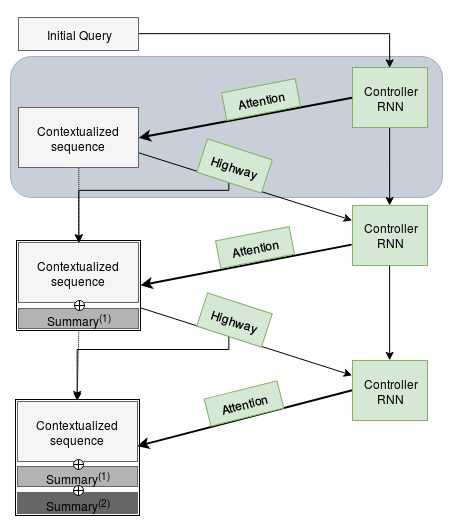}
\caption{The iterative recursive attention model (IRAM). Green-colored components share their parameters
with components of the same type. Highlighted in gray is one iteration of IRAM.}
\label{fig:attention}
\end{figure}

Fig.~\ref{fig:attention} shows the architecture of the iterative attention mechanism.
The mechanism uses a recurrent network, dubbed the \emph{controller}, to refine the
attention query throughout $T$ iterations. 

Inputs to the mechanism are an initial query $\hat{x}$ and a set of hidden states $H = [h_i, \ldots, h_N]$ constituting the
input sequence, both obtained from the encoding step.

As the controller, we use a gated recurrent unit (GRU) \cite{cho2014properties} cell. The input to the controller is the transformed result of the
previous query, while the hidden state is the previous query.

For the attention mechanism we use bilinear attention \cite{luong2015effective}:
\begin{equation}
\vec{a} = \mathit{softmax}(\vec{q} WH)
\label{eq:attn}
\end{equation}
\noindent where  $\vec{q}$ is the current query vector, $W$ a parameter of size $\mathbb{R}^{d_q \times d_h}$, while $d_q$ and $d_h$
are the dimensionalities of the query and the hidden state, respectively.

The attention weights are then used to compute the input summary in timestep $t$ as a linear combination of the hidden states:
\begin{equation}
\hat{s}^{(t)} = \sum_i^N a_i^{(t)} h_i
\end{equation}

As we intend to use $\hat{s}^{(t)}$ in the next iteration
of the attention mechanism, we need to allow the network the capacity to
discern between the new additions and original inputs. 
To this end, we use
a highway network \cite{srivastava2015highway}, which gives the model the option to pass subsets of
the summary as-is or transform them with a nonlinearity. If the summary is not transformed
with a nonlinearity, it ends up being merely a linear combination of the hidden states, and we gain no
information from adding it to the sequence.

The final input summary is thus
obtained as \mbox{$s^{(t)} = \mathit{Highway}(\hat{s}^{(t)})$} and added to the set of
hidden states $H = \{h_i, \ldots, h_n, s^{(1)}, \ldots, s^{(t)}\}$.

\subsection{Attention Regularization}

Ideally, we want the model to focus on different task-related aspects of the input in each iteration. 
However, the model is in no way incentivized to learn to propagate information through the summaries and can in principle focus on the same segment in each step.

To prevent this from happening -- and push the model to focus on different aspects of the input in every step -- we regularize it by minimizing the pairwise dot products between all iterations of attention:
\begin{equation}
L_{\mathit{attn}} = \frac{\gamma}{2T} \sum_{i\neq j} [A A^T]_{ij}
\end{equation}
\noindent where $\gamma$ is a hyperparameter determining the regularization strength and $A \in \mathbb{R}^{T\times N+T-1}$ is a matrix containing the attention weights generated in $T$ steps over $N$ inputs by the iterative attention mechanism. The matrix has $N+T-1$ columns to account for attention over $T-1$ added summaries, as the summary generated in the last iteration cannot be attended over. In each row $t$, the matrix has $T-t-1$ trailing zeroes, corresponding to summaries that are not yet available in iteration $t$. 

Concretely, the attention weight vector in row $t$ of the matrix $A$ consists of:

\begin{equation}
	A_t = \lbrack \overbrace{a_{1}, \dots, a_{N}}^{\text{Input sequence}}, \overbrace{a_{N+1}, \dots, a_{N+t-1}}^{\text{Summaries in $t^- < t$}}, 0, \dots \rbrack 
\end{equation}

\noindent resulting in each element $i,j$ of the regularization matrix $A A^T$ storing the dot product between attention weights in iterations $i,j$.
The regularization expression is a sum over all off-diagonal elements. The diagonal elements are dot products of attention weights in the same iteration so we ignore them. We scale by
$\frac{1}{2}$ to account for the symmetrical elements in $A^T A$ and by $\frac{1}{T}$ to account for the number of dot product comparisons.

We note that, while this regularization penalty does encourage the model to focus on different elements of the input sequence, there is still a trivial way for the model to minimize the penalty without learning a meaningful behavior. Since the attention weight over the summary in iteration $t$ is zero in all iterations $t^- < t$, the model can simply attend over any elements of the input sequence
in the first iteration, and afterwards propagate the information forward by fully attending only over the summary generated in the previous iteration. We will illustrate this behavior with 
concrete examples in Section~\ref{sec:experiments}.

\subsection{Vanilla Encoder}
\label{sec:vanilla}

For training, the inputs of the encoding phase are a sequence of words $x = [w_1, \ldots, w_N]$ and a
class label $y$. 
The encoder of the vanilla model maps the word indices to dense vector representations using pretrained GloVe vectors \cite{pennington2014glove}.
The sequence of word vectors is then fed as input to a bidirectional long-short term memory (BiLSTM) network \cite{hochreiter1997long}. The outputs of the BiLSTM are
used as the input sequence to the iterative attention step, while the cell state in the last timestep is used as the initial query.

\subsection{Full Encoder}
\label{sec:full}

\begin{figure}
\centering
\includegraphics[width=7.5cm]{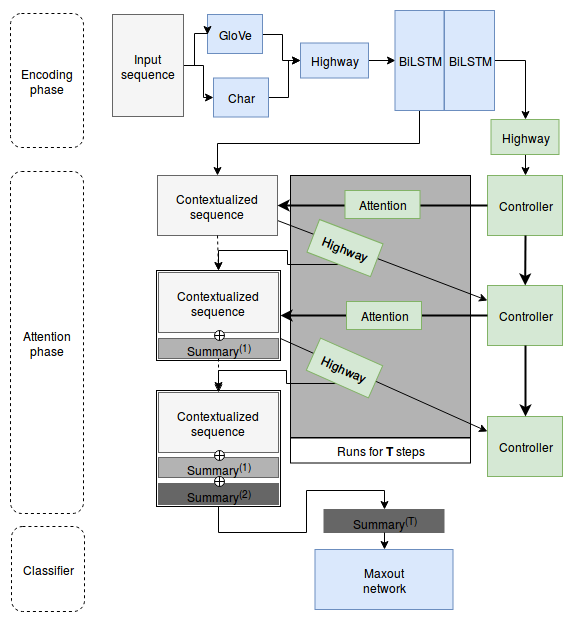}
\caption{The full version of the iterative recursive attention model (IRAM). Green-colored components share their parameters
with components of the same type; blue-colored components each have their
own parameters.}
\label{fig:model}
\end{figure}

There are three key differences between the full encoder and the vanilla encoder. The full encoder uses (1) character n-gram embeddings, (2) an additional highway network, whose task is
to fine-tune the word embeddings, and (3) an additional layer of BiLSTM, followed by a highway layer to construct the initial query. 
For extensions (1) and (2), we took inspiration from \newcite{mccann2017learned}, 
who also use both components. However, unlike \newcite{mccann2017learned}, who used a ReLU feedforward network to fine-tune the embeddings for the task, we use a highway
network, which we found performs better.

The pretrained character n-gram vectors obtained from \cite{hashimoto2016joint} are first
averaged over all character n-grams for a given word and then concatenated to the GloVe embedding.
Further on, before feeding the sequence of word
embeddings to a recurrent model, we use a two-layer highway network
\cite{srivastava2015highway} to fine-tune the embeddings for the task,
which is especially beneficial when the input vectors are kept fixed.

To contextualize the input sequence and produce an initial attention query, we
use a bidirectional long-short term memory (BiLSTM) network. We split the network conceptually into two parts: 
the lower $l_{\mathit{ctx}}$ layers are used to transform the input sequence of
word embeddings into a sequence of contextualized word representations, while
the upper $l_{\mathit{query}}$ layers are used to read and comprehend the now-transformed
sequence and capture its relevant aspects into a single vector. 
The rationale for the split is that recurrent
networks are often required to tackle two tasks at once: contextualize the input
and comprehend the whole sequence. Intuitively, the split should incite a
division of labor between the two parts of the network: contextualization
network only has to memorize the local information specific to each word (e.g., verb tense, noun gender) in
order to transform its representation, while comprehension
network needs to model aspects of meaning pertaining to the entire sequence (e.g., the overall sentiment of the sentence, locations of sentiment bearing phrases).

We use a single $(l_{\mathit{ctx}} + l_{\mathit{query}})$-layered BiLSTM, where we use 
the output of the $l_{\mathit{ctx}}$-th layer, while we use the cell state from
the last layer as the sequence representation $\hat{x}$.

Lastly, since the weights of the BiLSTM network are suited toward processing the input sequence rather than preparing the query vector, we add an additional highway layer designed
to fine-tune the sentence representation into the initial query.

\subsection{Classifier}

As input to the classifier, we use the summary vector obtained from the last
step of iterative attention $s^{(T)}$. This way we force the
network to propagate information through the attention steps, and also because 
the intermediate summaries do not contribute directly toward the
classification and hence need not have the same polarity.
The last summary vector is fed into a maxout network
\cite{goodfellow2013maxout} to obtain the class-conditional probabilities.

Fig.~\ref{fig:model} shows the full version of the iterative attention mechanism with all of the aforementioned components.

%% file: 4-Experiments.tex

\section{Experiments}
\label{sec:experiments}

\subsection{Datasets}

We test IRAM on two sentiment classification datasets. 
%
%
The first is the Stanford Sentiment Treebank (SST) 
\cite{socher2013recursive}, a dataset derived from movie reviews on Rotten Tomatoes and containing 11,855 sentences labeled into five classes at the sentence
level and at the level of each node in the constituency parse tree. 
 The binary version with the neutral class removed contains 56,400 instances, while the fine-grained version with scores ranging from 1 (very negative) to 5 (very positive) contains 94,200 text-sentiment pairs.
The second dataset is the Internet Movie Database (IMDb) 
\cite{maas2011learning}, containing 22,500 multi-sentence reviews extracted from 
positive and negative reviews. We truncate each sentence from this dataset to a maximum length of 200
tokens.

Firstly, we demonstrate and analyze how each component in the vanilla model contributes to the performance and interpretability. We then analyze the full model and evaluate it on the aforementioned datasets.



\subsection{Experimental Setup}

Unless stated otherwise, all weights are initialized from a Gaussian
distribution with zero mean and standard deviation of $0.01$. We use the Adam optimizer
\cite{kingma2014adam} with the AmsGrad modification \cite{reddi2018convergence}
and $\alpha = 0.0003$. We clip the global norm of the gradients to $1.0$ and
set weight decay to $0.00003$.

We use $300$-dimensional GloVe word embeddings trained on the Common Crawl
corpus and $100$-dimensional character embeddings. We follow the
recommendation of \newcite{mu2017all} and standardize the embeddings. Dropout
of $0.1$ is applied to the word embedding matrix.

For both datasets, we set $l_{\mathit{ctx}} = 2$ and $l_{\mathit{query}} =
1$. 
The highway network for fine-tuning the input embeddings has two layers, while
the ones fine-tuning the query and the summary have a single layer. All highway
networks' gate biases are initialized to $1$, as recommended by \newcite{srivastava2015highway},
as well as the biases of the LSTM forget gates.

The maxout network uses two $200$-dimensional layers with a pool size of $4$.

Throughout our experiments, we have experimented with selecting the
batch size from $\{32, 64\}$, dropout for the recurrent layers and the maxout
classifier from $\{0.1, 0.2, 0.3,0.4\}$, and the LSTM hidden state size from $\{400, 500,
1000\}$. The word and character n-gram vectors are kept fixed for SST but are
learned for the IMDb dataset. These parameters are optimized using cross-validation,
and the best configuration is ran on the test set. As IMDb has no official
validation set, we randomly select 10\% of the
dataset and use it for all of the experiments. The values of other
hyperparameters were selected through inexhaustive search.

\subsection{Analysis of the Vanilla Model}

The vanilla model defined in Section~\ref{sec:vanilla} has two main confounding
variables: strength (and presence) of attention regularization ($\gamma$) and the
number of iterations of the iterative recursive attention mechanism ($T$). We also would
like to examine the difference in performance of the vanilla IRAM compared to
some baseline sequence classifier. To this end, we implement a baseline model
without the attention mechanism -- a maxout classifier over the last hidden
state of the encoder BiLSTM.
To keep the running time of the experiments feasible, in this section we use only the binary SST dataset.

\paragraph{Effect of regularization.}

\begin{figure} 
\centering
\includegraphics[width=8cm]{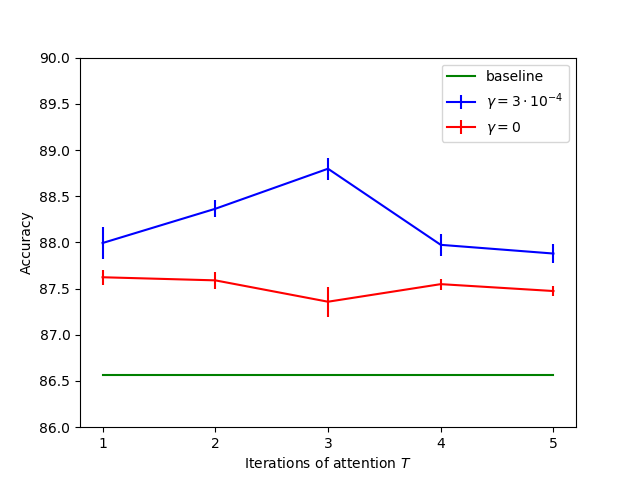}
\caption{The effect of regularization $\gamma$ across different values of $T$}
\label{fig:exist-gamma}
\end{figure}

\begin{figure}
\centering
\includegraphics[width=8cm]{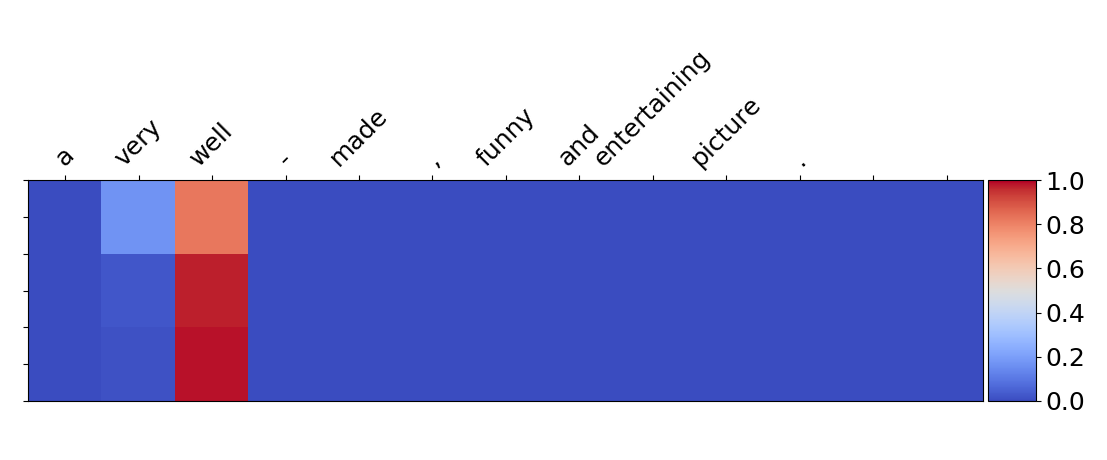}
\caption{Attention sample for $\gamma =0$ and $T=3$}
\label{fig:no-reg-attn}
\end{figure}

For each experiment in this round, we run every model three times with different random seeds and report the average results along with the standard deviations across the experiments.
In Fig.~\ref{fig:exist-gamma} we present the comparison between the performance of the vanilla model with and without regularization. A more telling sign of the different behavior between the models can be seen through inspecting attention weights. 

In Fig.~\ref{fig:no-reg-attn} we can see that the attention mechanism, when not regularized, fails to use its capacity and simply attends over the same element in each time-step. 
The last two columns, which contain the summaries from the first two steps of the iterative attention mechanism, have an attention weight of $0$, which means that the model does not pass any information through the summaries nor refine the query. This behavior initially prompted us to add the regularization penalty term.

Through inexhaustive search we isolated a critical range of values for $\gamma$, for which we perform a detailed analysis of performance. For this experiment, we fix $T=3$ as it 
has exhibited better performance for the vanilla model.

\begin{figure}
\centering
\includegraphics[width=8cm]{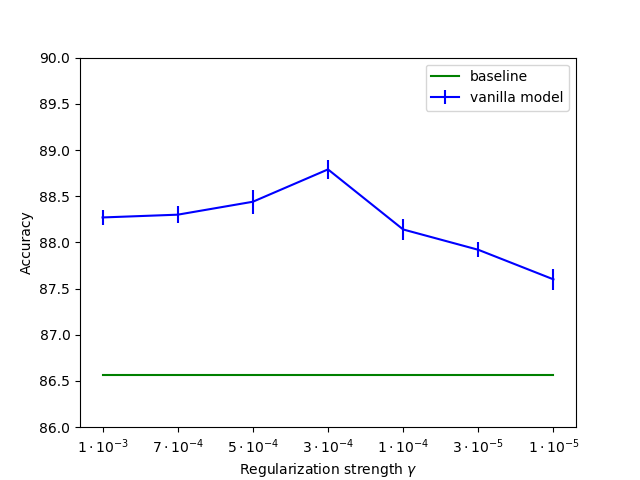}

\caption{Classification accuracy for different values of $\gamma$}
\label{fig:abl-gamma}
\end{figure}

\paragraph{Effect of the number of iterations.}

Apart from comparing the effect of the existence of regularization, in Fig.~\ref{fig:exist-gamma} we can also observe the effect of the number of timesteps $T$. Increasing $T$ beyond $3$ has a diminishing effect on classification performance, something which we find to be consistent for the IMDB dataset as well.

We attribute this decrease in performance to the fact that SST is relatively simple, containing at most two contrastive aspects in each sentence, making any additional steps unnecessary. While the model could in theory exploit the pass-through mechanism, we believe that this operation adds some noise to the final representations and in turn affects performance slightly.

\subsection{Analysis of the Full Model}

We now evaluate the full model. 
Table~\ref{table:res-sota} shows the accuracy scores of our best models (for $T=3$, $\gamma = 0.0003$) and other
state-of-the-art models on the test portions of the SST and IMDb datasets. Our model performs
competitively with the best results on SST and SST-5 datasets. It is important to note that our model does not use transfer learning apart from the pretrained word vectors, which is not the case for the competing models. 

\begin{table}
\centering
\small
\begin{tabular}{ |l|c| }
\hline
\multicolumn{2}{|c|}{SST} \\
\hline
 NSE \cite{munkhdalai2017neural} & 89.7 \\
\textbf{IRAM} & 90.1 \\
 BCN + CoVe \cite{mccann2017learned} & 90.3 \\ 
 bmLSTM \cite{radford2017learning} & \textbf{91.8}  \\ 
 \hline
 \multicolumn{2}{|c|}{SST-5}  \\
 \hline
 \textbf{IRAM} & 53.7 \\
 BCN + CoVe \cite{mccann2017learned} & 53.7 \\
 BCN + ELMo \cite{peters2018deep} & \textbf{54.7} \\
  \hline

 \multicolumn{2}{|c|}{IMDb}  \\
 \hline
 \textbf{IRAM} & 91.2 \\
 TRNN \cite{dieng2016topicrnn} & 93.8 \\
 oh-LSTM \cite{johnson2016supervised}& \textbf{94.1} \\
 Virtual \cite{miyato2016adversarial} & \textbf{94.1}\\
 
\hline
\end{tabular}
\caption{Classification accuracy on the test sets}
\label{table:res-sota}
\end{table}

\paragraph{Ablation study of encoder components.} 

As mentioned in Section~\ref{sec:full}, through adding various components to the model we introduced a number of confounders.
In order to determine the effect of each of the added components on the overall score, we evaluate the performance of the full model on
the binary SST dataset
with the remaining hyperparameters fixed and one of the components removed in isolation.

\begin{table}
\centering
\small
\begin{tabular}{ |l|c| }
\hline
\textbf{Removed component} & \textbf{Accuracy} \\
\hline
Full model & \textbf{90.1} \\
Vanilla model & 88.7 \\
\hline
-- char n-grams & 89.3\\
-- query fine-tune & 89.8\\
-- embedding fine-tune & 89.3\\
\hline

\end{tabular}
\caption{Effect of removing components on performance}
\label{table:res-ablate}
\end{table}

\subsection{Visualizing Attention}

\begin{figure}[t]
\centering
\begin{subfigure}[b]{0.9\columnwidth}
\includegraphics[width=7.5cm]{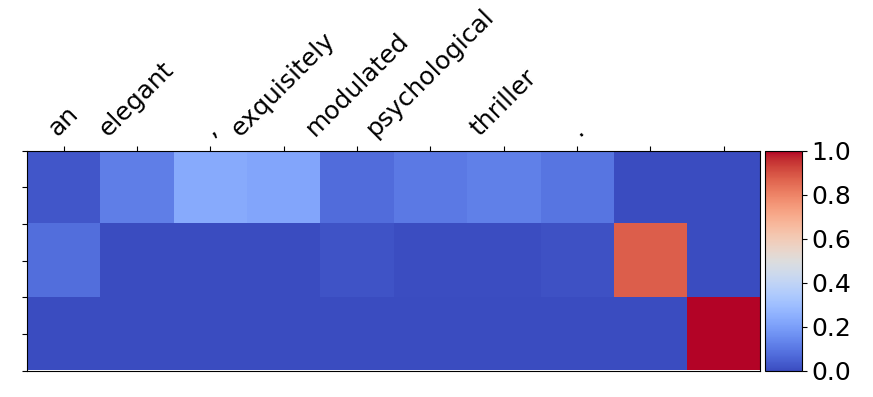}
\caption{Simple unipolar sentence}
\label{viz:simple}
\end{subfigure}
\begin{subfigure}[b]{0.9\columnwidth}
\includegraphics[width=7.5cm]{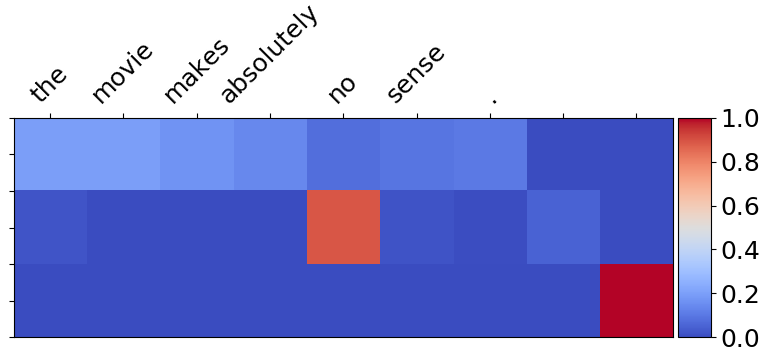}
\caption{Sentence with a negation}
\label{viz:simple-neg}
\end{subfigure}
\begin{subfigure}[b]{0.9\columnwidth}
\includegraphics[width=7.5cm]{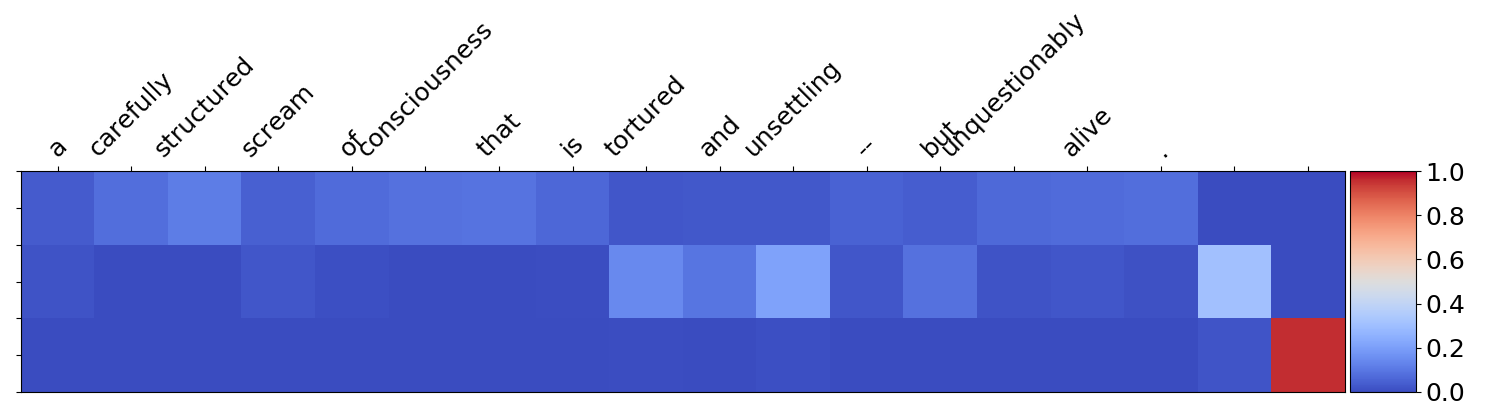}
\caption{Contrastive multipolar sentence}
\label{viz:long}
\end{subfigure}
\caption{Visualization of attention across sentence words (horizontal) and $T$=3 time steps (vertical). 
The last $T$-1 columns contain the attention weights over the result of the previous attentive query.}
\end{figure}

To gain an intuition about the working of IRAM, we visually analyzed its attention mechanism on a number of sentences from our dataset. We limit ourselves to examples from the test set of the SST dataset
as the length of examples is manageable for visualization. We isolate three specific cases where the attention mechanism 
demonstrates interesting results: (1) simple unipolar sentences, 
(2) sentences with negations, and (3) multipolar sentences. 

The least interesting case is the unipolar, as the attention mechanism often
does not need multiple iterations. Fig.~\ref{viz:simple} shows the attention mechanism 
simply propagating information, since sentiment classification is straightforward and does not
require multiple attention steps. This can be seen from most of the attention weight in the 
second and third steps being on the columns corresponding to the summaries.

The more interesting cases  are sentences involving negations and modifiers. 
Fig.~\ref{viz:simple-neg} shows the handling of negation: attention is initially on all words except on the negator. 
In the second step, the mechanism combines the output of the first step
with the negation. We interpret this as flipping the sentiment -- the model
cannot rely solely on recognizing a negative word, and has to account for what
that word negates through a functional dependence. These examples highlight one of the drawbacks of 
recurrent networks which we aim to alleviate. In case a standard attention mechanism is applied to a
sentence containing a negator, the hidden representation of the negator has to scale or negate the 
intensity of an expression. Our model has the capacity to process such sequences iteratively, first
constructing the representation of an expression, which is then adjusted by the nonlinear transformation
and simpler to combine with the negator in the next step.

Lastly, 
Fig.~\ref{viz:long} shows a contrastive multipolar sentence, where the model
in the first step focuses on positive words, and then combines the negative
words (\emph{tortured}, \emph{unsettling}) with the results of the first step.
In such cases, the model succeeds to isolate the contrasting aspects of the sentence
and attends to them in different iterations of the model, alleviating the burden of
simultaneously representing the positive and negative aspects. After both contrastive
representations have been formed, the model has the capacity to \emph{weigh} them one against
other and compute the final representation.

%% file: 5-Conclusion.tex
\section{Conclusion}
\label{sec:conclusion}

The proposed iterative recursive attention model (IRAM) has the
capacity to construct representations of the input sequence in a recursive 
fashion, making inference more interpretable. 
We demonstrated that the model can learn to focus
on various task-relevant parts of the input, and can propagate the information in
a meaningful way to handle the more difficult cases. On the sentiment analysis task, the model performs
comparable to the state of the art. Our next goals will be to try to use the iterative attention mechanism
to extract tree-like sentence structures akin to constituency parse trees, 
evaluate the model on more complex datasets as well as extend the model to support
an adaptive number of iterative steps.